\documentclass[letterpaper, 10 pt, conference]{ieeeconf}  

\IEEEoverridecommandlockouts                              

\usepackage{amsmath}
\usepackage{graphicx}
\usepackage{multirow}
\usepackage{multicol}
\usepackage{todonotes}
\usepackage{subfigure}
\usepackage{caption}
\usepackage{float}

\usepackage{array}

\newcommand{\be}{\begin{equation}}
\newcommand{\ee}{\end{equation}}
\newcommand{\beq}{\begin{equation}}
\newcommand{\eeq}{\end{equation}}
\newcommand{\bed}{\begin{displaymath}}
\newcommand{\eed}{\end{displaymath}}
\newcommand{\beqa}{\begin{eqnarray}}
\newcommand{\eeqa}{\end{eqnarray}}
\newcommand{\beqann}{\begin{eqnarray*}}
\newcommand{\eeqann}{\end{eqnarray*}}
\newcommand{\bseq}{\begin{subequations}}
\newcommand{\eseq}{\end{subequations}}

\newcommand{\ba}{\begin{array}}
\newcommand{\ea}{\end{array}}

\newcommand{\1}{{\bf 1}}

\newcommand{\real}{\hbox{I \kern -.5em R}}

\title{\LARGE \bf
Kinova Gen3~Lite manipulator inverse kinematics: optimal polynomial solution*
}

\author{Hamed Montazer Zohour$^{1}$, Bruno Belzile$^{1}$ and David St-Onge$^{1}$
\thanks{*This work was supported by NSERC Discovery Grant (RGPIN-2020-06121).}
\thanks{$^{1}$École de technologie supérieure, Montréal, Department of Mechanical Engineering,
        {\tt\small hamed.montazer-zohour.1@ens.etsmtl.ca,  bruno.belzile.1@ens.etsmtl.ca, david.st-onge@etsmtl.ca}}%
}

\begin{document}

\maketitle
\thispagestyle{empty}
\pagestyle{empty}

\begin{abstract}

A polynomial solution to the inverse kinematic problem of the Kinova Gen3 Lite robot is proposed in this paper. This serial robot is based on a 6R kinematic chain and is not wrist-partitioned. We first start from the forward kinematics equation providing the position and orientation of the end-effector, finally, the univariate polynomial equation is given as a function of the first joint variable $\theta_{1}$. The remaining joint variables are computed by back substitution. Thus, an unique set of joint position is obtain for each root of the univariate equation. Numerical examples, simulated in ROS (Robot Operating System), are given to validate the results, which are compared to the coordinates obtained with MoveIt! and with the actual robot. A procedure to choose an optimum posture of the robot is also proposed.

\end{abstract}

\section{INTRODUCTION}
\label{s:intro}
Robotic manipulators can be found in a wide range of industrial applications, namely to conduct pick-and-place operations. To be able to automate these tasks, a symbolic solution to the inverse kinematics problem (IKP) is a powerful tool for control. The vast majority of commercial manipulators with 5 or 6 revolute joints (commonly referred to as 5R and 6R) are said to be wrist-partitioned (such as the Kuka KR15 and ABB IRB). These manipulators lead to a closed-form solution to their IKP. Specific conditions must be met so the inverse kinematics of 6R serial manipulators can be decoupled, i.e., conditions on architecture parameters for which the orientation and positioning problem can be solved separately~\cite{pieper_kinematics_1968}. It often leads to wrist joint analogue to a spherical configuration. This condense configuration of the wrist is complex to fully enclose, for instance to prevent any finger of a user to be trapped or pinched. The Kinova Gen3 Lite described in this work, is a 6R serial robot for collaborative operations, i.e. to achieve tasks close to a user. Its design optimise safety and the reachable work space, but falls into the category of non wrist partitioned manipulators.

As respectively shown by Pimrose~\cite{primrose_input-output_1986} and Lee et al.~\cite{lee_complete_1991}, a general 6R robotic manipulator has a maximum number of 16 different solutions to its IKP for a given end-effector pose. A polynomial degree 16 is the lowest possible that can be obtained for an univariate polynomial equation describing the kinematics of the robot. Polynomial solutions for different manipulators can be found in the literature~\cite{manseur_robot_1989,gosselin_polynomial_2014} with similar methodologies as the one described in this work. Considering 16th degree polynomial equations are prone to numerical ill-conditioning as well as the possibility of {\em polynomial degeneration} with roots yielding an angle of $\pi$, Angeles and Zanganeh proposed a semi-graphical solution to the inverse kinematics of a general 6R serial manipulator~\cite{angeles_semigraphical_1993}. However, these techniques do not apply to non wrist partitioned manipulators.

Numerical methods have also been applied by several researchers~\cite{chen_numerical_1999,aghajarian_inverse_2011,duleba_comparison_2013}, but these are commonly known to be prone to instability near singular postures. Moreover, they only give one possible solution, which may not be optimum. Several algorithms, including the ones proposed by Mavroidis et al.~\cite{mavroidis_inverse_1994}, Husty et al.~\cite{husty_new_2007} and Qiao et al.~\cite{qiao_inverse_2010}, can be found in the literature to find the 16th degress univariate polynomial equation for a 6R robotic manipulator, the latter notably using double quaternions.

\begin{figure} 
\centering
	\includegraphics[scale=0.65]{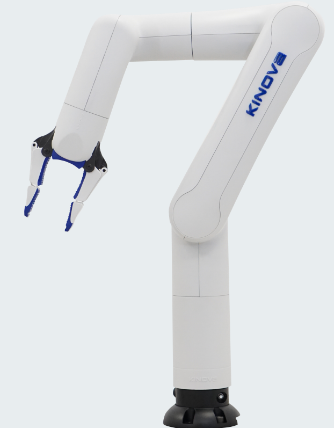}
	\caption{The 6R serial Kinova Gen3 Lite robot}
	\label{fig:kinova1}
\end{figure}

Among the 16 solutions to the IKP, a wide range of methodologies has been proposed to select the best posture. As these solutions are theoretical, one must first discard the one that cannot be implemented: non real roots, exceeding joint limits or resulting into a self-colliding posture. From there, simple algorithms such as the minimization of the amount of joint rotation can easily be implemented. Task-dependent optimization can also be used for certain applications and performance indices based on the kinematics (eg. kinetostatic conditioning index) and the stiffness (eg. deformation evaluation index) of the robot~\cite{lin_posture_2017}. In this work, a task-dependent procedure is proposed to select one optimum solution in order to avoid occlusion from a top-view camera while conducting pick-and-place operations.


We leverage a similar methodology to the one introduced by Gosselin and Liu~\cite{gosselin_polynomial_2014} to obtain an univariate polynomial equation to solve the IKP of the Kinova Gen3 Lite robot, shown in Fig.~\ref{fig:kinova1}. All joint angles are computed by back substitution. Configuration examples are given and we compare with the solutions obtained with a numerical IKP solver. Finally, a methodology to select a single solution is proposed and validated experimentally. The Python script used to solve the IKP, compute all real solutions (postures) and select the best following our application is made public.


\section{System under Study}
\label{s:system}
The Kinova Gen3 Lite is a serial manipulator with six revolute joints each having limited rotation and a two-finger gripper as the end-effector. The Denavit-Hartenberg (DH) parameters of this robot are given in Table~\ref{table:DH}, where the non-zero parameters are identified. With the parameters in this table, it is clear that this robot is not wrist-partitioned since $b_{5}\neq0$. Thus, well-known methodologies to find the decoupled solution of the IKP cannot be used.

\begin{table}[h!]
	\caption{DH parameters of the Kinova Gen3 Lite}
	\begin{center}
		\label{table:DH}
    \begin{tabular}{c||c|c|c|c|c|c} 
	i & 1 & 2 & 3 & 4 & 5 & 6\\ [0.5ex] 
	\hline \hline
	$a_i$ & 0 & $a_{2}$ & 0 & 0 & 0 & 0\\
	\hline
	$b_i$ & $b_{1}$ & $b_{2}$ & $b_3$ & $b_4$ & $b_5$ & $b_6$\\
	\hline
	$\alpha_i$ & $\pi/2$ & $\pi$ & $\pi/2$ & $\pi/2$ & $\pi/2$ & 0\\
		\end{tabular}
	\end{center}
\end{table}

\begin{figure} 
\centering
	\includegraphics[scale=0.33]{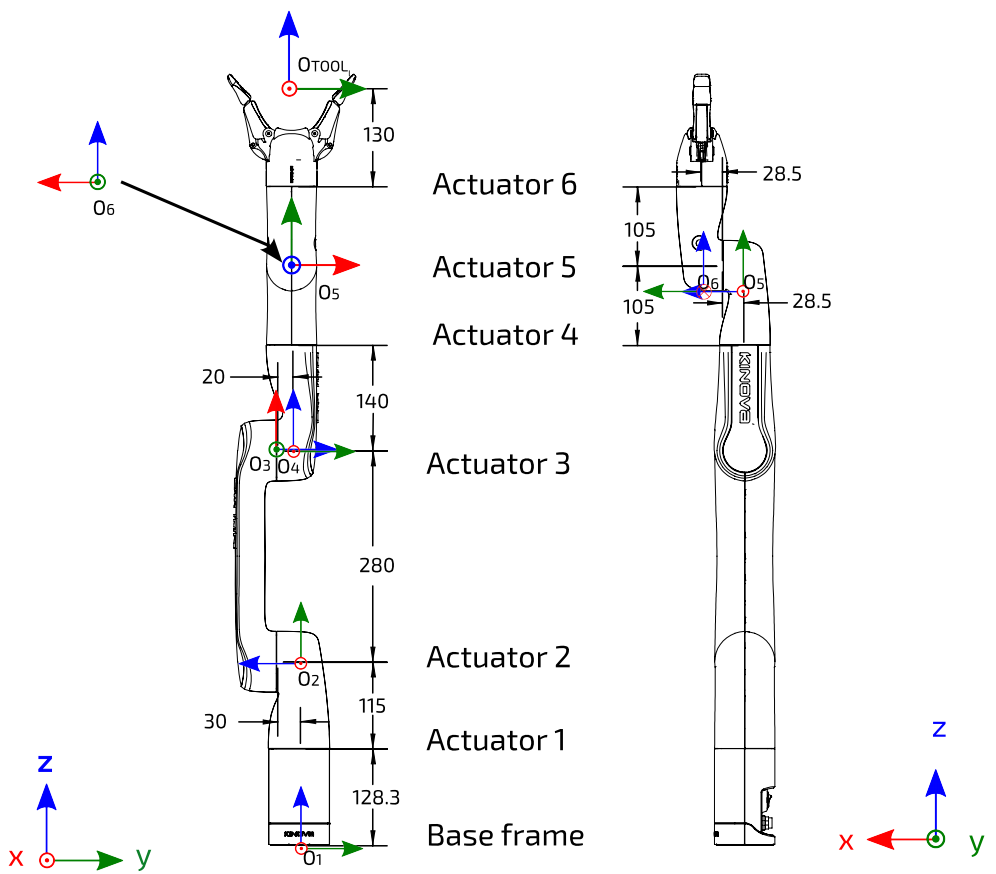}
	\caption{DH frames for each joint with the link dimensions (extracted from the manipulator user manual).}
	\label{fig:DH}
\end{figure}

As shown in Fig.~\ref{fig:DH}, a DH reference frame is attached to each link. It should be noted that these frames are not necessary located at the joints. the rotation matrices $Q_{i}$ and the position vectors $a_{i}$ relating the successive reference frames defined on each of the links of the robot~\cite{mavroidis_inverse_1994} can be written as

\begin{equation}
    \mathbf{Q}_{i}=
\begin{bmatrix}
	\cos\theta_{i} & -\cos\alpha_{i}\sin\theta_{i} & \sin\alpha_{i}\sin\theta_{i}\\
	\sin\theta_{i} & \cos\alpha_{i}\cos\theta_{i} & -\sin\alpha_{i}\cos\theta_{i}\\
	0 & \sin\alpha_{i} & \cos\alpha_{i}
\end{bmatrix}
\end{equation}
and
\begin{equation}
	\mathbf{a}_{i}=
\begin{bmatrix}
    a_{i}\cos\theta_{i} &
    a_{i}\sin\theta_{i} &
    b_{i}
\end{bmatrix}^T
\end{equation}
where the rotation matrix $\mathbf{Q}_{i}$ transforms frame $i$ into frame $(i+1)$ and vector $a_{i}$ connects the origin of frame $i$ to the origin of frame $(i+1)$. The joint variables are noted $\theta_{i}$ while $a_{i}$, $b_{i}$ and $\alpha_{i}$ are the DH parameters representing the geometry of the Kinova Gen3 Lite. The end-effector is located at the origin of frame 7, which is defined by the 3-dimensional vector $\mathbf{p}$. The orientation of the end-effector is given by the rotation matrix from frame 1 to frame 7, noted $\mathbf{Q}$.

\section{Forward Kinematics}
\label{s:FK}
The forward kinematic problem (FKP), i.e. the Cartesian position $\mathbf{p}$ and orientation matrix of the tool $\mathbf{Q}$, are straight forward and can be written as
\bseq
\be
    \mathbf{p}=\sum_{i=0}^5 \left(\left(\prod_{j=0}^i \mathbf{Q}_j\right)\mathbf{a}_{i+1}\right)
    \label{e:pos_EE}
\ee
and
\be
    \mathbf{Q}=\prod_{i=1}^6\mathbf{Q}_i
    \label{e:orientation_EE}
\ee
\eseq
where $\mathbf{Q}_0$ is the $3\times3$ identity matrix. 

\section{Inverse Kinematics}
\label{s:IK}
The first step toward obtaining a symbolic solution to the IKP of the Kinova Gen3 Lite is to reduce the number of unknowns, currently six for the six joints position $\{\theta_i\}$, to only one, therefore reducing the problem to a univariate polynomial equation that can be solved. Knowing these variables appear inside trigonometric functions $\sin\theta_i$ and $\cos\theta_i$, by finding expressions for these two and substituting them in the trigonometric identity $\sin^2\theta_i+\cos^2\theta_i=1$, we can readily reduce the number of unknowns.

First, to this aim, we need to compute the vector $\mathbf{r}$, connecting the origin of frame 1 to the origin of frame 6, which can be written similarly to Eq.~(\ref{e:pos_EE}) as
\be
    \mathbf{r}=\sum_{i=0}^4 \left(\left(\prod_{j=0}^i \mathbf{Q}_j\right)\mathbf{a}_{i+1}\right)
    \label{e:vector_r}
\ee
It is noted that vector $\mathbf{r}$ is independent of $\theta_6$ , so by premultiplying this equation by $\mathbf{Q}_1^T$ and isolating all expressions independent of $\theta_1$ on the righthand side, we have a set of nine scalar equations. Among them, two stand out as only being function of $\theta_1$, $\theta_2$ and $\theta_{(3-2)}$:
\begin{align}
    r_{1}c_{1}+r_{2}s_{1}&=a_{2}c_{2}+b_{5}c_{(3-2)}s_{4}+b_{4}s_{(3-2)}
    \label{e:implicit_c2}\\
    r_{3}-b_{1}&=a_{2}s_{2}-b_{5}s_{(3-2)}s_{4}+b_{4}c_{(3-2)}
    \label{e:implicit_s2}
\end{align}
where $r_i$ is the $i$\textsuperscript{th} component of $\mathbf{r}$, $s_i$ and $c_i$ respectively for $\sin\theta_i$ and $\cos\theta_i$, where $c_{(i-j)}$ and $s_{(i-j)}$ stand respectively for $\cos(\theta_i -\theta_j)$  and $\sin(\theta_i - \theta_j)$. Another equation also stands out after premultiplying Eq.~(\ref{e:vector_r}) by $\mathbf{Q}_1^T$ and will be needed later in the derivation:
\be
    r_{1}s_{1}-r_{2}c_{1}=b_{2}-b_{3}+b_{5}c_{4}.
    \label{e:implicit_c4}
\ee
It can be rewritten to obtain an explicit expression of $c_4$:
\be
    c_{4} =\frac{1}{b_{5}}(r_{1}s_{1}-r_{2}c_{1}+b_{3}-b_{2})
    \label{e:explicit_c4}
\ee
We are now able to solve Eqs.~(\ref{e:implicit_c2}-\ref{e:implicit_s2}) for $s_2$ and $c_2$. Substituting the results in $s_2^2+c_2^2=1$ as mentioned above, we obtain
\bseq
\be
    B_{1}s_{(3-2)}+B_{2}c_{(3-2)}+B_{3} = 0
    \label{e:implicit_s(3-2)c(3-2)-1}
\ee
where
\begin{align}
    B_{1}&=2(r_{3}-b_{1})b_{5}s_{4}-2(r_{1}c_{1}+r_{2}s_{1})b_{4}\\
    B_{2}&=-2b_{4}(r_{3}-b_{1})-2b_{5}s_{4}(r_{1}c_{1}+r_{2}s_{1})\\
    B_{3}&=(r_{3}-b_{1})^{2} +(r_{1}c_{1}+r_{2}s_{1})^{2} +b_{4}^{2}+b_{5}^{2}s_{4}^{2}-a_{2}^{2}
\end{align}
\eseq

Having a first equation expressed as a function of $s_{(3-2)}$ and $c_{(3-2)}$, a second one is needed to be able to use the same trigonometric identity and compute $s_{(3-2)}^2+c_{(3-2)}^2=1$.

Rotation matrices $\mathbf{Q}$ are orthogonal matrices ($\mathbf{Q}_i\mathbf{Q}_i^T = \mathbf{I}$), Eq.~(\ref{e:orientation_EE}) can be recast into the following form:
\be
    \mathbf{Q}_{4}\mathbf{Q}_{5}\mathbf{Q}_{6}=\mathbf{Q}_{3}^{T}\mathbf{Q}_{2}^{T}\mathbf{Q}_{1}^{T}\mathbf{Q}
    \label{e:Q4Q5Q6}
\ee
This equation gives us a system of nine scalar equations. However, only five are relevant: the ones defining the first two components of the last row and the three components of the last column of the resulting matrices. On the one hand, the former can be used to obtain explicit expressions of $c_6$ and $s_6$:
\bseq
\begin{align}
    c_{6}&=\frac{q_{11}c_{1}s_{(3-2)}+q_{21}s_{1}s_{(3-2)}+q_{31}c_{(3-2)}}{s_{5}} \label{e:explicit_c6}\\
    s_{6}&=\frac{q_{12}c_{1}s_{(3-2)}+q_{22}s_{1}s_{(3-2)}+q_{32}c_{(3-2)}}{-s_{5}} \label{e:explicit_s6}
\end{align}
\eseq
These two equations will be useful later in the paper. On the other hand, the components of the last column are not a function of $\theta_6$, because the latter corresponds to a rotation of the last joint about the z-axis of the end-effector. Therefore, the last column, defining a unit vector parallel to this axis, must be independent of $\theta_6$. With this column, we obtain the following scalar equations:
\bseq
\begin{align}
    c_{4}s_{5}=&q_{13}c_{1}c_{(3-2)}+q_{23}s_{1}c_{(3-2)}-q_{33}s_{(3-2)}
    \\
    s_{4}s_{5}=&-q_{13}s_{1}+q_{23}c_{1}
    \\
    -c_{5}=&q_{13}c_{1}s_{(3-2)}+q_{23}s_{1}s_{(3-2)}+q_{33}c_{(3-2)}
\end{align}
\eseq
By casting these three equations in array form with dyalitic elimination, we have
\bseq
\be
    \mathbf{M} \mathbf{k}_{5} = \mathbf{0}
    \label{e:Mk5}
\ee
where $\mathbf{0}$ is a three-dimensional zero vector and
\be
    \mathbf{M}=
    \begin{bmatrix}
	0 & -c_{4} & m_{13}\\
    0 & -s_{4} & m_{23}\\
    1 & 0 & m_{33}
    \end{bmatrix},\quad
    \mathbf{k}_{5} = \begin{bmatrix}c_{5}\\s_{5}\\1\end{bmatrix}
\ee
with, after some simplifications, 
\begin{align} 
    m_{13} &= (q_{13}c_{1}+q_{23}s_{1})c_{(3-2)}-q_{33}s_{(3-2)}
    \\
    m_{23} &= (-q_{13}s_{1}+q_{23}c_{1})
    \\
    m_{3} &= (q_{13}c_{1}+q_{23}s_{1})s_{(3-2)}+q_{33}c_{(3-2)}
    \\
    s_i&\equiv\sin\theta_i,\quad c_i\equiv\cos\theta_i
    \\
    s_{(i-j)}&\equiv\sin(\theta_i-\theta_j),\quad c_{(i-j)}\equiv\cos(\theta_i-\theta_j)
\end{align}
\eseq
In the above expressions, $q_{ij}$ is the (i,j)\textsuperscript{th} component of the end-effector orientation matrix $\mathbf{Q}$. It can be seen that $\mathbf{M}$, an homogeneous matrix, in Eq.~(\ref{e:Mk5}) is singular, as vector $\mathbf{k}_5$ cannot vanish. Therefore, we have
\bseq
\be
    \mathrm{det}(\mathbf{M}) = A_{1} s_{(3-2)}+A_{2} c_{(3-2)}+A_{3} = 0
\label{e:implicit_s(3-2)c(3-2)-2}
\ee
where
\begin{align}
    A_{1}&=-q_{33}s_{4}\\
    A_{2}&=q_{13}c_{1}s_{4}+q_{23}s_{1}s_{4}\\
    A_{3}&=q_{13}c_{4}s_{1}-q_{23}c_{1}c_{4}
\end{align}
\eseq
Equations~(\ref{e:implicit_s(3-2)c(3-2)-1} \& \ref{e:implicit_s(3-2)c(3-2)-2}) can now be solved for $s_{(3-2)}$ and $c_{(3-2)}$, and substituted in $s^2_{(3-2)}+c^2_{(3-2)}=1$, yielding
\bseq
\begin{align}
    c_{(3-2)}&=(A_3 B_1-B_3 A_1)/(B_2 A_1 - A_2 B_1) \label{e:explicit_c(3-2)}\\
    s_{(3-2)}&=(A_3 B_2-B_3 A_2)/(B_2 A_1 - A_2 B_1) \label{e:explicit_s(3-2)}
\end{align}
and, finally,
\be
    \begin{split}
            (A_{2}B_{3}-A_{3}B_{2})^{2}+(A_{3}B_{1}-A_{1}B_{3})^{2}\\
            -(A_{1}B_{2}-A_{2}B_{1})^{2} = 0
    \end{split}
    \label{e:AB}
\ee
\eseq

\begin{figure*}%
\centering
\subfigure[Solution \#2]{%
\includegraphics[height=2.5in]{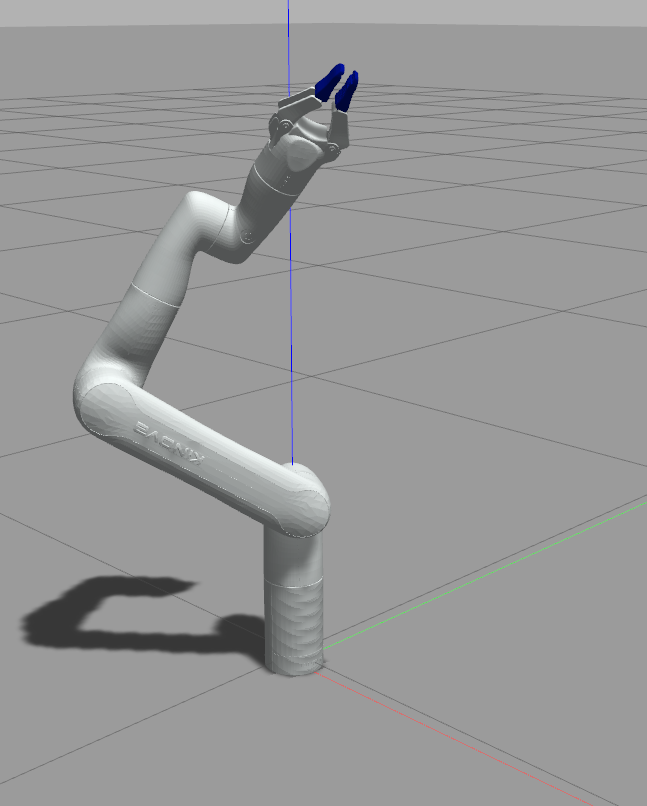}}
\hspace{3cm}
\subfigure[Solution \#4]{%
\includegraphics[height=2.5in]{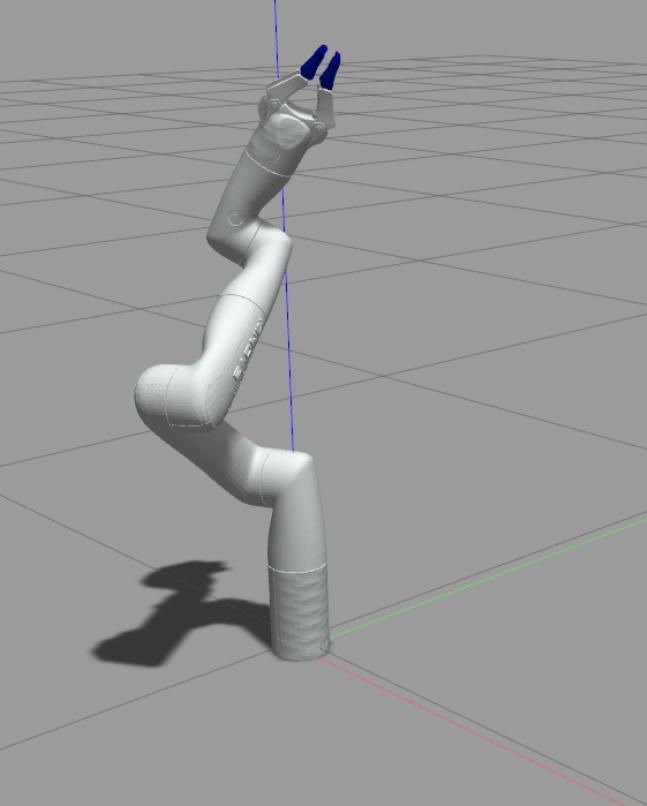}}
\caption{Excerpt of possible postures from example \#1. The end-effector pose is the same, but the manipulator configurations (joint angles) differ.}
\label{fig:ex1}
\end{figure*}

Having eliminated all expressions of $\theta_2$ and $\theta_3$ with the procedure above, Eq.~(\ref{e:AB}) is only a function of $\theta_1$ and $\theta_4$, bringing us closer to our objective of finding a univariate polynomial equation. Equation~(\ref{e:AB}) can be factorized as a function of powers of $c_4$ and $s_4$, giving us 
\be
    \begin{split}
        F_{1}c_{4}^{6}+F_{2}c_{4}^{5}+F_{3}c_{4}^{4}+F_{4}c_{4}^{3}s_{4}+F_{5}c_{4}^{3}+F_{6}c_{4}^{2}s_{4}\\
        +F_{7}c_{4}^{2}+F_{8}c_{4}s_{4}+F_{9}c_{4}+F_{10}s_{4}+F_{11} = 0
    \end{split}
    \label{e:F}
\ee
where the coefficients $F_{i},\quad i=1,\dots,11$ are solely dependent of $\theta_{1}$. With Eq.~(\ref{e:explicit_c4}), Eq.~(\ref{e:F}) becomes
\bseq
\be
    Vs_{4}+W = 0
    \label{e:implicit_s4}
\ee
with
\begin{align}
    \begin{split}
    V &=v_{1}c_{1}^{3}+v_{2}c_{1}^{2}s_{1}+v_{3}c_{1}^{2}+v_{4}c_{1}s_{1}+v_{5}c_{1}\\
    &+v_{6}s_{1}+v_{7}
    \end{split}
    \\
    \begin{split}
    W &= w_{1}c_{1}^{4}+w_{2}c_{1}^{3}s_{1}+w_{3}c_{1}^{3}+w_{4}c_{1}^{2}s_{1}+w_{5}c_{1}^{2}\\
    &+w_{6}c_{1}s_{1}+w_{7}c_{1}+w_{8}s_{1}+w_{9}
    \end{split}
\end{align}
\eseq
where $v_{i}$ and $w_{i}$ are only functions of the DH parameters and the orientation $\mathbf{Q}$ and position $\mathbf{p}$ of the tool. The above equation can be solved for $s_4$, then substituted, with Eq.~(\ref{e:explicit_c4}) in $s^2_4+c^2_4=1$. The resulting univariate equation is
\be
    b_{5}^{2}W^{2}+[(r_{1}s_{1}-r_{2}c_{1}+b_{3}-b_{2})^{2}-b_{5}^{2}]V^{2} = 0
    \label{e:implicit_c1s1}
\ee
Equation~(\ref{e:implicit_c1s1}) is one of degree 8 in terms of $c_1$ and of degree 1 in terms of $s_1$. Then, using the Weierstrass substitution ($c_{1}=({1-T_{1}^{2}})/({1+T_{1}^{2}})$ and $s_{1}=(2T_{1})/({1+T_{1}^{2}})$), Eq.~(\ref{e:implicit_c1s1}) is finally transformed into a polynomial in $T_1=\tan(\theta_1/2)$:
\be
    \sum^{16}_{i=0}  E_{i}T_{1}^{i} = 0
    \label{e:univariate_polynomial}
\ee
where $\{E_{i}\}$ are functions of the DH parameters, the position and the orientation of the Kinova Gen3 Lite. The roots of this univariate polynomial can then be computed to obtain $T_1$, then leading to the values of $\theta_1$.

\section{Back substitution}
\label{s:backsubstitution}

As mentioned above, the roots of Eq.~(\ref{e:univariate_polynomial}) can be computed to find all theoretically values of $\theta_1$. Some of these solutions may be complex numbers and some can be duplicates. For control, only the real roots can be considered. Using a subset of the equations presented in Section~\ref{s:IK}, it is possible to compute all other joint angles for each real solution. For all remaining joint angles, a single trigonometric function is needed, i.e.:
\be
    \theta_i=\mathrm{arctan2}(s_i,c_i)
    \label{e:atan2}
\ee
The equation numbers for expressions of $s_i$ and $c_i$ are given in Table~\ref{t:backsubstitution}. The back substitution procedure must be conducted following the order from left to right, top to bottom presented in this table, starting with $c_4$. Finally, $\theta_3$ is easily computed from $(\theta_3-\theta_2)$ and $\theta_2$.
\begin{table}[h!]
	\caption{Back substitution}
	\begin{center}
		\label{t:backsubstitution}
    \begin{tabular}{c||c|c} 
	i & $c_i$ & $s_i$ \\ [0.5ex] 
	\hline \hline
	$\theta_4$          & Eq.~(\ref{e:explicit_c4}) & Eq.~(\ref{e:implicit_s4}) \\ \hline $\theta_3-\theta_2$ &Eq.~(\ref{e:explicit_c(3-2)}) & Eq.~(\ref{e:explicit_s(3-2)})  \\ \hline
	$\theta_5$          & Eq.~(\ref{e:Mk5}) (last row) & Eq.~(\ref{e:Mk5}) (second row) \\ \hline
	$\theta_2$          & Eq.~(\ref{e:implicit_c2}) & Eq.~(\ref{e:implicit_s2}) \\ \hline
	$\theta_6$          & Eq.~(\ref{e:explicit_c6}) & Eq.~(\ref{e:explicit_s6})
		\end{tabular}
	\end{center}
\end{table}

\section{Special Cases}
\label{s:special}
Like the majority of similar algorithms, some special cases must be considered. The special cases considered here are similar to those pointed out by Gosselin and Liu~\cite{gosselin_polynomial_2014} for another manipulator. First, it is possible that coefficient $V$ in Eq.~(\ref{e:implicit_s4}) becomes equal to zero. Since, according to the procedure detailed in the previous section, both $s_4$ and $c_4$ are required, the value of $\theta_4$ cannot be computed with Eq.~(\ref{e:atan2}). Instead, $arccos$ must be used, and two values of $\theta_4$ for a single $\theta_1$ will be obtained. Of course, since the total number of solutions cannot exceed 16, some will be repeated.

Another possible special case arise when $(B_2A_1-A_2B_1)$ is equal to zero. Thereby, Eqs.~(\ref{e:explicit_c(3-2)} \& \ref{e:explicit_s(3-2)}) cannot be computed. Instead, Eqs.~(\ref{e:implicit_s(3-2)c(3-2)-1} \& \ref{e:implicit_s(3-2)c(3-2)-2}) are solved for $\theta_{(3-2)}$ with the Weierstrass substitution previously mentioned, leading to two solutions for $\theta_{(3-2)}$ for a single $\theta_1$. As always, no more than 16 unique sets of joint angles can be obtained, which means there will be some repeated solutions again.

\begin{table}[h!]
	\caption{Numerical parameters of the Kinova Gen3 Lite}
	\begin{center}
		\label{t:parameters}
    \begin{tabular} 
    {m{0.75cm}||m{0.75cm}|m{0.75cm}|m{0.75cm}|m{0.75cm}|m{0.75cm}|m{0.75cm}}
	i & 1 & 2 & 3 & 4 & 5 & 6\\ [0.5ex] 
	\hline \hline
	$a_i$ & 0 & 0.28 & 0 & 0 & 0 & 0\\
	\hline
	$b_i$ & 0.2433 & 0.03 & 0.02 & 0.245 & 0.057 & 0.235\\
	\hline
	$\alpha_i$ & $\pi/2$ & $\pi$ & $\pi/2$ & $\pi/2$ & $\pi/2$ & 0\\
    	\hline
	Lower limit &$-154^{\circ}$&$ -150^{\circ}$&$-150^{\circ}$&$-149^{\circ}$&$-145^{\circ}$&$-149^{\circ}$\\[0.3ex]
    	\hline
	Upper limit&$+154^{\circ}$&$+150^{\circ}$&$+150^{\circ}$&$+149^{\circ}$&$+145^{\circ}$&$+149^{\circ}$\\[0.3ex]
		\end{tabular}
	\end{center}
\end{table}

\section{Examples and Validation}
\label{s:examples}

This section presents and discuss two examples to illustrate the IKP presented above. A Python script was written to process all the equations and is publicly available online~\cite{montazer_kinova_2020}. The results are validated with ROS-Gazebo simulation, as shown in Fig~\ref{fig:ex1}. It should be noted that while some solutions may be theoretically possible, they are not feasible in practice because of the mechanical limits of the joints. The numerical values of the DH parameters and the joints' limitations are given in Table~\ref{t:parameters}.

Finally, the roll-pitch-yaw angles are used to give the orientation of the end-effector. Incidentally, the orientation matrix $\mathbf{Q}$ is defined as
\bseq
\be
    \mathbf{Q}\equiv
    \begin{bmatrix}
       \mathbf{q}_1 & \mathbf{q}_2 & \mathbf{q}_3\end{bmatrix}
    \label{e:rpy}
\ee
with
\begin{align}
    \mathbf{q}_1=&
    \begin{bmatrix}
        \cos\psi\cos\theta\\
        \sin\psi\cos\theta\\
        -\sin\theta
    \end{bmatrix}
\\
    \mathbf{q}_2=&
    \begin{bmatrix}
        -\sin\psi\cos\phi+\cos\psi\sin\theta\sin\phi\\
        \cos\psi\cos\phi+\sin\psi\sin\theta\sin\phi\\
        \cos\theta\sin\phi
    \end{bmatrix}
\\
    \mathbf{q}_3=&
    \begin{bmatrix}
        \sin\psi\sin\phi+\cos\psi\sin\theta\cos\phi\\
        -\cos\psi\sin\phi+\sin\psi\sin\theta\cos\phi\\
        \cos\theta\cos\phi
    \end{bmatrix}
\end{align}
\eseq
where $\phi$, $\theta$ and $\psi$ are the roll, pitch and yaw angles, respectively.

\subsection{Example \#1}
\label{ss:ex1}

\begin{figure}[H]
\centering
	\includegraphics[scale=0.75]{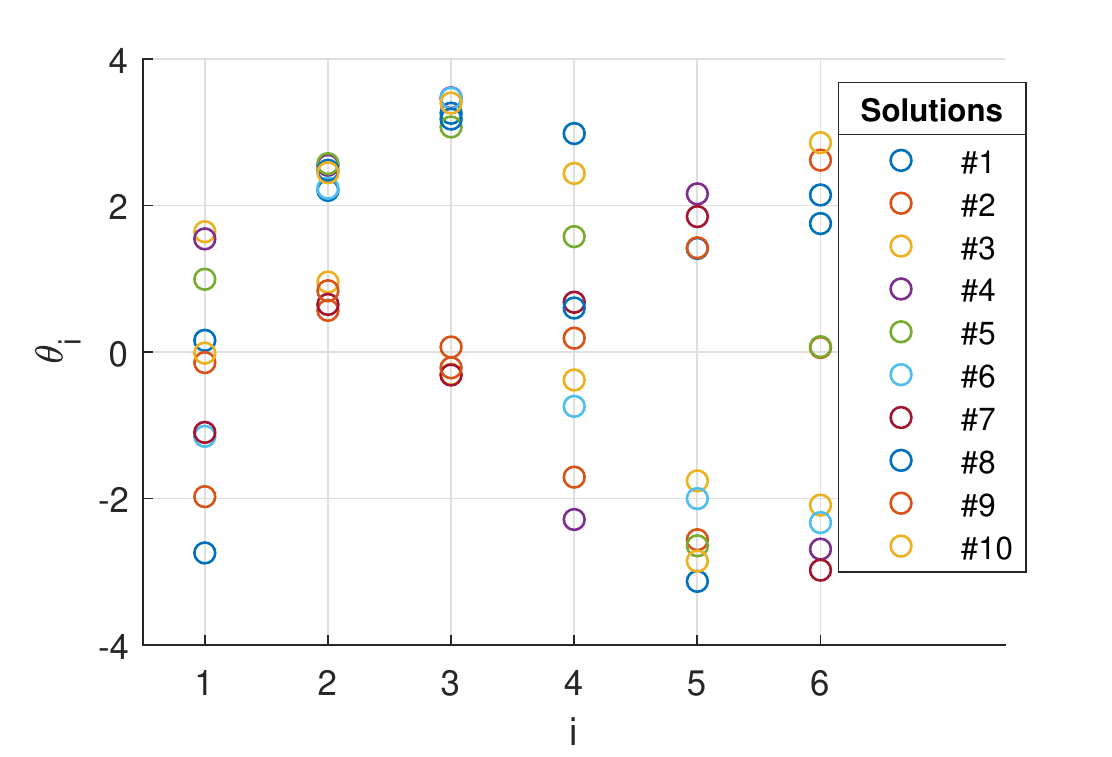}
	\caption{Solutions to example \#1.  Each of the six joints angle ($-\pi>\theta_i<\pi$) are shown for all 10 solutions computed.}
	\label{fig:sol1}
\end{figure}

\begin{table}[H]
	\caption{Example \#1}
	\begin{center}
		\label{t:ex1}
    \begin{tabular}{m{0.9cm}||m{0.60cm}|m{0.65cm}|m{0.60cm}|m{0.75cm}|m{0.50cm}|m{0.75cm}} 
	\multirow{2}{0.9cm}{Joint space} & $\theta_1$ & $\theta_2$ & $\theta_3$ & $\theta_4$ & $\theta_5$ & $\theta_6$\\ [0.5ex] 
	
	 & 1 & 1 & 1.5 & 0 & 0.5 & -1.5\\
	\hline \hline
	\multirow{2}{0.9cm}{Cartesian space} & $x$~[m] & $y$~[m] & $z$~[m] & $\phi$ & $\theta$ & $\psi$\\
	
	& 0.119& -0.04& 0.763& -0.527& 0.47& -0.759\\
		\end{tabular}
	\end{center}
\end{table}
For this example, the end-effector position $p$ and orientation $\mathbf{Q}$ were first obtained from a set of joint coordinates with the forward kinematics (see Section~\ref{s:FK}). This initial set of joint coordinates and the corresponding position and orientation of the end-effector are detailed in Table~\ref{t:ex1}. The simulation results are depicted in Fig.~\ref{fig:ex1}. The obtained solutions are shown in Fig.~\ref{fig:sol1}. It should be noted that 10 solutions were initially found by solving the IKP. However, only 6 were within the joint limitations, detailed in Table~\ref{t:sol1}. 


\begin{table}[h!]
	\caption{Feasible solutions to example \#1}
	\begin{center}
		\label{t:sol1}
    \begin{tabular}{m{0.72cm}||m{0.77cm}|m{0.77cm}|m{0.77cm}|m{0.77cm}|m{0.77cm}|m{0.75cm}} 
	Sol. & $\theta_1$ & $\theta_2$ & $\theta_3$ & $\theta_4$ & $\theta_5$ & $\theta_6$\\ [0.5ex] 
	\hline \hline
	4 & 1.544 & 0.979 & 1.900 & 2.425 & -0.982 & 2.021\\
	\hline 
	5 & 0.993 & 1.001 & 1.502 & 0.005 & 0.496 & -1.499\\
	\hline 
	6 & -1.151 & 0.665 & 1.895 & -2.313 & 1.140 & 2.383\\
	\hline
	7 & -1.098 & -0.921 & -1.885 & -0.891 & -1.029 & 1.734\\
	\hline
	8 & 0.160 & 0.910 & 1.609 & -0.970 & 0.010 & 0.183\\
	\hline
	9 & -0.145 & -0.735 & -1.786 & -1.382 & -1.718 & 1.049\\
	\hline
	\hline
	MoveIt! & 1.54 & 0.98 & 1.90 & 2.40 & -0.98 & 2.00]\\
	\hline
	Robot & 1.59 & 1.00 & 1.93 & 2.39 & -1.00 & 2.01\\
		\end{tabular}
	\end{center}
\end{table}
We also included in Table~\ref{t:sol1} the numerical solutions obtained with ROS MoveIt! IK package and with the actual robot controller. It can be found among the solutions obtained with the procedure detailed in Section~\ref{s:IK}.


\begin{figure}[H]
\centering
	\includegraphics[scale=0.85]{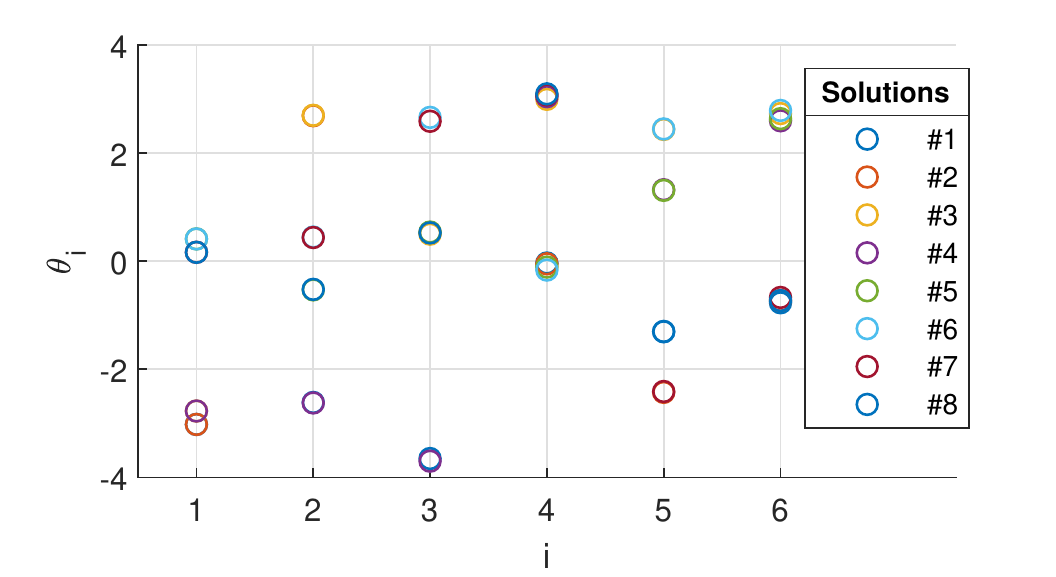}
	\caption{Solutions to example \#2. Each of the six joints angle ($-\pi>\theta_i<\pi$) are shown for all 8 solutions computed.}
	\label{fig:sol2}
\end{figure}

\begin{figure*}%
\centering
\subfigure[Solution \#5]{%
\includegraphics[height=1.8in]{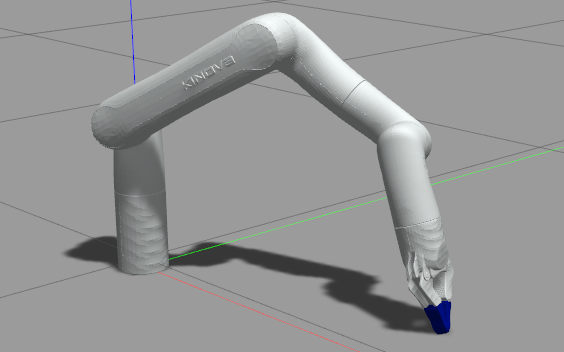}}
\hspace{2cm}
\subfigure[Solution \#8]{%
\includegraphics[height=1.8in]{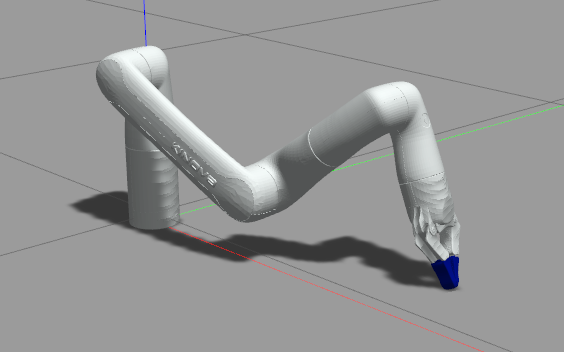}}
\caption{Excerpt of possible postures from example \#2. The end-effector reach to a pick-and-place location on the table.}
\label{fig:ex2}
\vspace{-1em}
\end{figure*}

\subsection{Example \#2}
\label{ss:ex2}

\begin{table}[H]
	\caption{Example \#2}
	\begin{center}
		\label{t:ex2}
    \begin{tabular}{m{0.9cm}||m{0.55cm}|m{0.55cm}|m{0.75cm}|m{0.555cm}|m{0.75cm}|m{0.55cm}} 
	\hline \hline
	\multirow{2}{0.9cm}{Cartesian space} & $x$~[m] & $y$~[m] & $z$~[m] & $\phi$ & $\theta$ & $\psi$\\
	
	& 0.503& 0.122& -0.002& 3.077& -0.254& 0.256\\
		\end{tabular}
	\end{center}
\end{table}
In this example, we simulated a pick-and-place task. To be able to grasp the object, the position and orientation of the end-effector were first determined, as detailed in Table~\ref{t:ex2}. Then the IKP script was used leading to the solutions illustrated in Fig.~\ref{fig:sol2}. Those within the joint limitations are detailed in Table~\ref{t:sol2}, as well as the numerical solution obtained with ROS MoveIt! IK and the actual robot numerical IK controller. Two of the solutions are depicted in Fig.~\ref{fig:ex2} and will be used in the next section to illustrate the selection of the optimal posture.

\begin{table}[H]
	\caption{Feasible solutions to example \#2}
	\begin{center}
		\label{t:sol2}
    \begin{tabular}{m{0.72cm}||m{0.75cm}|m{0.77cm}|m{0.77cm}|m{0.77cm}|m{0.77cm}|m{0.77cm}} 
	Sol. & $\theta_1$ & $\theta_2$ & $\theta_3$ & $\theta_4$ & $\theta_5$ & $\theta_6$\\ [0.5ex] 
	\hline \hline
	5 & 0.415 & -2.010 & -1.030 & -1.678 & -1.829 & -1.444\\
	\hline 
	6 & 0.414 & -1.122 & 1.092 & -1.733 & -0.692 & -1.292\\
	\hline 
	7 & 0.166 & -1.131 & 1.021 & 1.508 & 0.732 & 1.530\\
	\hline
	8 & 0.166 & -2.091 & -1.045 &  1.527 & 1.837 & 1.472\\
	\hline
	\hline
	MoveIt! & 0.40 & -0.87 & 1.10 & -1.55 & -0.96 & -1.05\\
	\hline
	Robot & 0.45 & -2.20 & -1.19 & -1.74 & -1.76 & -1.32\\
		\end{tabular}
	\end{center}
\end{table}

\section{Optimal Posture}
\label{s:optimum}

\begin{figure}[H]
\centering
	\includegraphics[scale=0.25]{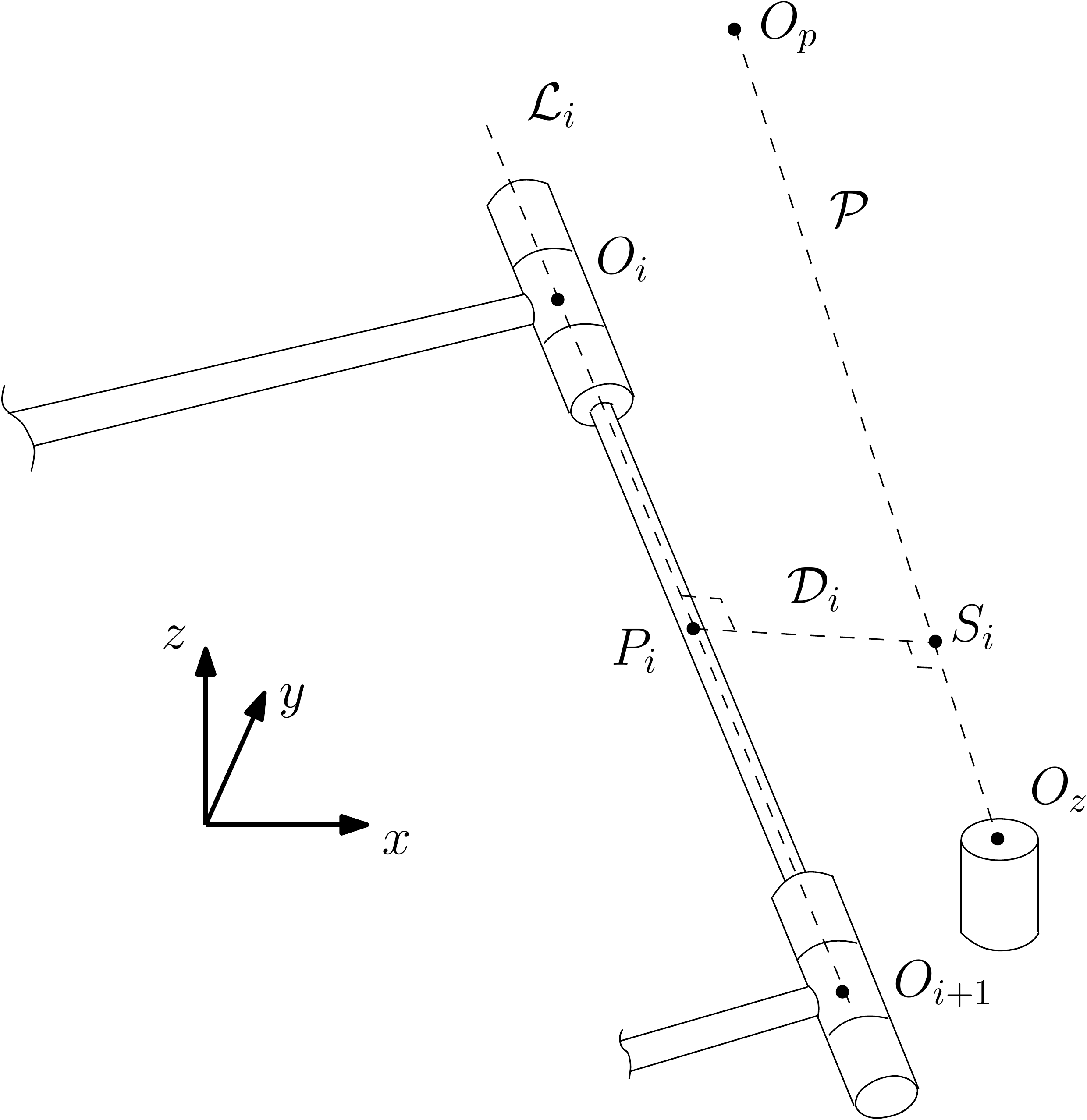}
	\caption{Schematic of the distance ($D_i$) computed between the arm link and the line of sight to the objects.}
	\label{fig:selection}
\end{figure}

Except for some particular cases, more than one solution emerge from solving the IKP. Thus a strategy is required to select the best fitted solution; a single set of joint angles. 
A wide range of procedures can be used to select that optimal solution following the task (such as manipulating fragile objects) and the application context (such as low energy requirements). Our approach targets pick-and-place tasks relying on a top-view camera, positionned above the table work space. The optimisation criterion is to maximise the field of view. This can be extend to several pick-and-place operations. The objective is thus to avoid the manipulator interfering with the camera's line of sight with the objects on the table. To this aim, the shortest distance between all links and the line of sight to all objects must be maximized, as depicted in Fig.~\ref{fig:selection}.


\begin{figure*}%
\centering
\subfigure[Solution \#5]{%
\includegraphics[height=2in]{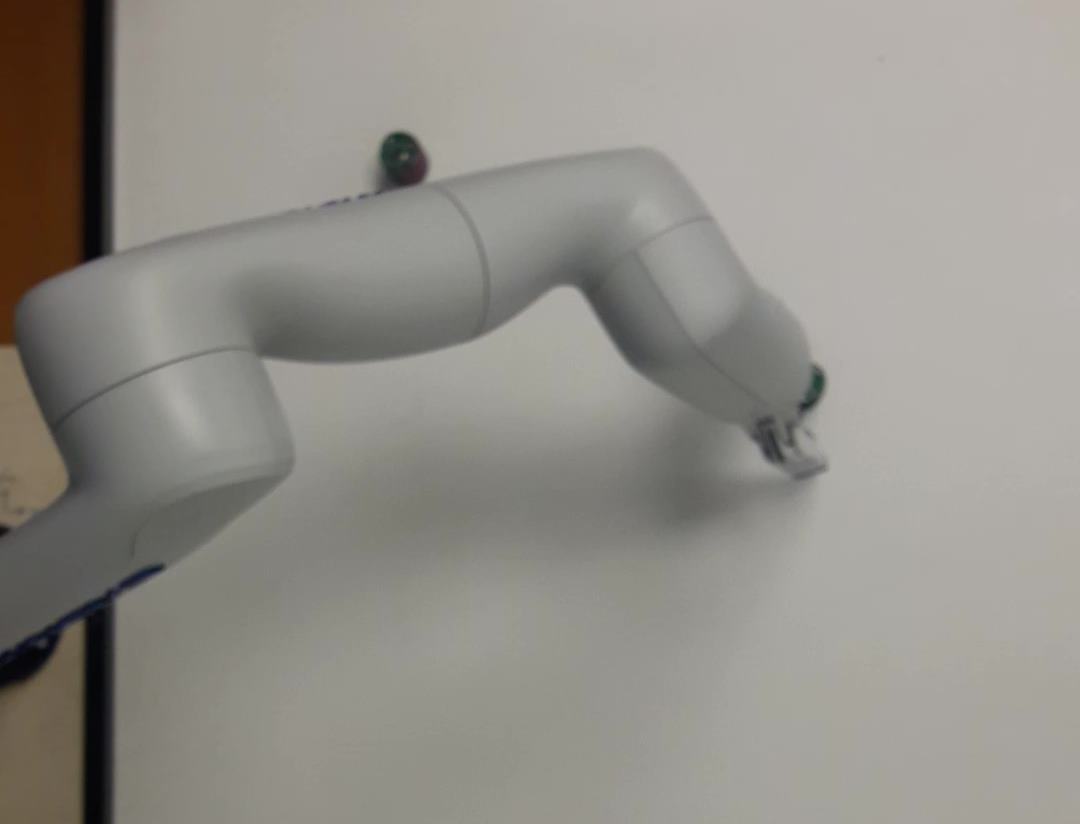}}%
\hspace{2cm}
\subfigure[Solution \#8]{%
\includegraphics[height=2in]{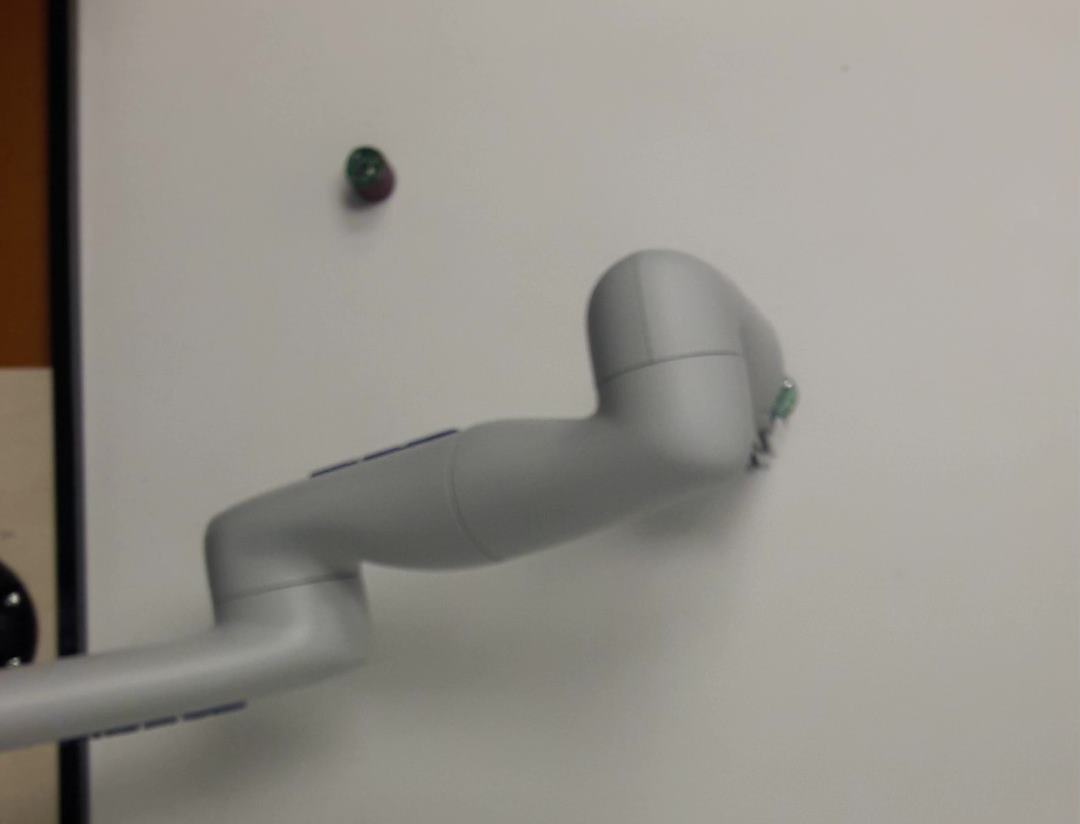}}%
\caption{Two configurations of the arm for the same object picking task. On the left, the other object is almost hidden, while the right solution has a lot more margin.}
\label{fig:experimental}
\vspace{-1em}
\end{figure*}

First, the position of a point along the straight line $\cal{P}$ from the projector, located at $O_p$, to an object, located at $O_z$, is defined as
\be
    \mathbf{s}_i = \mathbf{O}_p+\Delta_{p,i}(\mathbf{O}_z-\mathbf{O}_p)
\ee
where $\Delta_{p,i}$ is a factor defining where along the line this point is located. Moreover, the Cartesian coordinates of points $S_i$, $O_p$ and $O_z$ are, respectively, arrayed in vectors $\mathbf{S}_i$, $\mathbf{O}_p$ and $\mathbf{O}_z$. Similarly, the position of a point $P_i$ along the line ${\cal{L}}_i$ can be defined for any given link of the manipulator, i.e.
\be
    \mathbf{p}_i = \mathbf{O}_i+\Delta_i(\mathbf{O}_{i+1}-\mathbf{O}_i),\quad i=1,\dots,6
\ee
where $\mathbf{O}_i$ and $\Delta_i$ are, respectively, the Cartesian coordinates of the $i$\textsuperscript{th}~joint and a factor defining where along this link this point is located. If these two points are the closest pair along their respective lines, a unit vector, orthogonal to ${\cal L}_i$ and $\cal P$, thus parallel to ${\cal D}_i$, can be defined as
\be
    \mathbf{v}_i=\frac{(\mathbf{O}_z-\mathbf{O}_p)\times(\mathbf{O}_{i+1}-\mathbf{O}_i)}{||(\mathbf{O}_z-\mathbf{O}_p)\times(\mathbf{O}_{i+1}-\mathbf{O}_i)||}
\ee
With these three vectors, a close loop equation can be formulated, i.e.
\be
    \mathbf{s}_i=\mathbf{p}_i+\Delta_{d,i}\mathbf{v}_i
\ee
where $\Delta_{d,i}$ is the shortest distance between ${\cal L}_i$ and $\cal P$. A set of three linear equations with three unknowns, $\Delta_i$, $\Delta_{p,i}$ and $\Delta_{d,i}$, is thus obtained and can easily be solved.

The value of these three unknowns obtained, the risk of occlusion for an object on the table can now be computed. Indeed, the shortest distance between the robot and $\overline{O_p O_z}$, namely $\mathrm{min}(\Delta_{d,1},\dots,\Delta_{d,6})$, for a prescribed end-effector position and orientation must be a large as possible. Of course, if point $P_i$ for a robot posture and a given link is not located within the limits of the latter, the corresponding $\Delta_{d,i}$ should be disregarded. It is the case, for instance, when $O_p$, $O_i$ and $O_{i+1}$ are aligned. Instead, the closest distance between a line ($\overline{O_p O_z}$) and a point (the corresponding link end) should be computed. This is done with the following equations:
\bseq
\begin{align}
    \Delta_{d,i} =& \frac{||(\mathbf{O}_p-\mathbf{O}_i)\times (\mathbf{O}_z-\mathbf{O}_p)||}{||\mathbf{O}_z-\mathbf{O}_p||},\quad\textrm{ if }\Delta_i<0    
    \\
    \Delta_{d,i} =& \frac{||(\mathbf{O}_p-\mathbf{O}_{i+1})\times (\mathbf{O}_z-\mathbf{O}_p)||}{||\mathbf{O}_z-\mathbf{O}_p||},\quad\textrm{ if }\Delta_i>1 
\end{align}
\eseq

With the postures presented in Table~\ref{t:sol1}, solution \#8, depicted in Fig.~\ref{fig:ex2}(b), is the one selected with this algorithm for $O_p=[0.329\quad0\quad1]^T$ and $O_z=[0.25\quad0.25\quad-0.002]^T$. The smallest distance between the robot and the line of sight is, in this case, 0.1723~m. Moreover, this test was validated experimentally, as shown in Fig.~\ref{fig:experimental}. The photos are taken from the camera located at $O_p$, showing clearly that solution \#8 is significantly better than solution \#4 with respect to the occlusion risk for the object located at $O_z$ (top left corner).

\section{Conclusion}
\label{s:conclusion}
In this letter, the inverse kinematic problem of the Kinova Gen3 Lite robot was studied. It was solved by finding a univariate polynomial equation to find all possible values of one angle, $\theta_1$, then finding the corresponding values of the other joint angular positions by back substitution. The Python script used to compute the solutions to the IKP is now public. Several examples were given and compared to the solutions obtained with ROS MoveIt! IK and the real robot controller for validation. Finally, a procedure to select the optimal solution in order to minimize the risk of occlusion while performing a pick-and-place task was proposed.








\bibliographystyle{ieeetran}
\bibliography{references}

\end{document}